*Article*

# FreeTransformSR: Efficient Lightweight Image Super-Resolution via Free Low-Rank Learnable Transform

**Hongji Li [1,2], Yun-Hui Li [2,*], Jianing Wang [2], Xiaodong Wang [2]**

[1] Changchun Institute of Optics, Fine Mechanics and Physics, Changchun, Jilin 130033, China
[2] University of the Chinese Academy of Sciences, Beijing 100049, China

**Abstract**

Single image super-resolution aims to reconstruct high-resolution images from low-resolution inputs. This paper proposes FreeTransformSR, a novel lightweight super-resolution network based on a channel-wise free low-rank learnable transform. The transform learns task-adaptive basis functions in a data-driven manner, enabling adaptive feature modulation with minimal parameter overhead. To further enhance high-frequency detail recovery, we introduce a local feature modulation branch that complements transform-domain processing with depthwise convolution. In addition, a soft complexity adaptive module dynamically fuses the outputs of local convolution and window self-attention branches through a lightweight gating network, adaptively adjusting the fusion ratio based on regional texture characteristics. An adaptive intensity modulation strategy is also incorporated to adjust transform-domain response strength at the sample level, enabling the network to dynamically adjust processing intensity according to input features. Extensive experiments on five benchmark datasets demonstrate that FreeTransformSR achieves competitive PSNR/SSIM performance with significantly fewer parameters and FLOPs. Specifically, FreeTransformSR achieves 32.41 dB on BSD100 (×2, 634K parameters) and 27.00 dB on Urban100 (×4, 653K parameters), while delivering faster inference speed than competing methods, making it well-suited for deployment in resource-constrained scenarios.

**Keywords:** Image super-resolution; Learnable transform; Channel-wise transform; Adaptive modulation; Local feature modulation; Soft complexity gating.

## 1. Introduction

Single Image Super-Resolution (SISR) is a classic task in computer vision and image processing, aiming to recover high-resolution (HR) images from their low-resolution (LR) counterparts. This technique has a wide range of applications, including medical imaging, digital photography, and reducing server deployment costs in streaming media transmission.

Since convolutional neural networks (CNNs) were first applied to single image super-resolution [1], a large number of CNN-based algorithms [2,3,4,5,6,7] have been proposed, continuously advancing the technical challenge of reconstructing HR images from LR inputs. However, convolution operations are inherently limited to local feature extraction and struggle to capture global dependencies. To overcome this, some CNN-based methods [4,7] construct extremely deep networks with complex structures to enlarge the receptive field. Although this improves reconstruction quality, it inevitably incurs substantial computational costs, severely limiting practical deployment.

In recent years, following the great success of Transformers in natural language processing, they have been widely applied to various high-level computer vision tasks [8,9,10,11,12,13], achieving remarkable results. The key to Transformer-based methods lies in the self-attention mechanism, which efficiently captures long-range dependencies. Leveraging the powerful feature modeling potential of Transformers, they have also attracted significant research attention in low-level computer vision tasks [14,15,16,17,18], including image super-resolution. The HAT model [15] experimentally demonstrates that fully utilizing more global information can effectively improve image reconstruction quality.

However, to alleviate the prohibitively high computational complexity of global self-attention, existing methods typically divide large images into multiple local regions for separate processing. While this strategy improves the computational efficiency of Transformer models and enables finer local feature extraction, it still suffers from several limitations. SwinIR [8] partitions images into content-agnostic fixed local windows, failing to fully exploit similar token features at distant positions, ultimately leading to suboptimal reconstruction quality. Axial stripe attention [15] expands the receptive field in a cross-shaped pattern, but this approach likewise lacks content-adaptive design, easily introducing substantial irrelevant interfering information.

To address these shortcomings, some studies have explored clustering-based solutions. For instance, SPIN [16] employs a soft K-means-based token clustering algorithm [22], using cluster centers as proxy features between queries and keys in the attention mechanism to facilitate long-range information propagation. However, SPIN has two major limitations. First, relying solely on cluster centers for long-range information propagation is a coarse approximation, which fails to precisely capture fine-grained long-range dependencies. Second, the algorithm requires iterative feature clustering during inference. This significantly slows down the model and hinders the deployment of lightweight applications. The ATD model [11] introduces an auxiliary dictionary to learn priors from training data and classifies tokens using the dictionary to achieve more accurate token grouping. However, ATD requires running multiple attention modules concurrently to improve accuracy, substantially increasing computational overhead and making it unsuitable for lightweight scenarios.

Addressing the limitations of existing methods, we propose FreeTransformSR, a novel efficient lightweight super-resolution network. As shown in Figure 1, the overall architecture employs learnable transform blocks and soft complexity adaptive blocks as core building units, constructing the backbone through stacked residual groups to achieve high-quality super-resolution reconstruction while strictly controlling parameter count and computational cost. The core innovations of this network are threefold.

First, at the feature transformation level, existing frequency-domain methods such as WavTrans [20] and CiaoSR [21] rely on predefined fixed basis transforms (e.g., discrete wavelet transform or Fourier transform), which are difficult to adaptively adjust according to image content. In contrast, we design a data-driven channel-wise learnable transform. Here, the term "Free" signifies that the transform basis is unconstrained by any predefined mathematical functions (like sine waves in Fourier). Instead, it learns a pair of trainable 2D transformation matrices (row transform and column transform) for each feature channel independently. Parameterized in the form of "identity mapping + low-rank residual" ($T = I + \Delta$ , with $\Delta$ initialized to zero), this transform maps features to a learnable domain for per-frequency modulation and then recovers spatial structure through transposed back-mapping. This offers stronger data adaptability compared to fixed-basis methods, while maintaining a computational complexity of only $O(C^2 \cdot N)$ , far lower than global self-attention's $O(N^2 \cdot C)$ .

Second, at the feature fusion level, we design a soft complexity adaptive module that dynamically fuses the outputs of a local convolution branch and a window self-attention branch through a lightweight gating network. The gate computes a continuous fusion coefficient based on input features, enabling the model to adaptively select an appropriate fusion ratio according to regional image characteristics, rather than making a hard binary choice between smooth and textured regions.

Third, at the network training level, we introduce an adaptive intensity modulation (scale gate) strategy. This module extracts global response features from input samples through a lightweight gating network, generating sample-level modulation coefficients to dynamically adjust the response strength of transform-domain processing. Specifically, global average pooling aggregates spatial information across channels, and a 1×1 convolution then linearly compresses the multi-channel responses into a unified scalar, applying a consistent scaling magnitude to all channels. The modulation coefficients are automatically guided by the loss function without explicit rule constraints—the network autonomously determines the degree of amplification or suppression for different inputs based on the statistical characteristics of the training data.

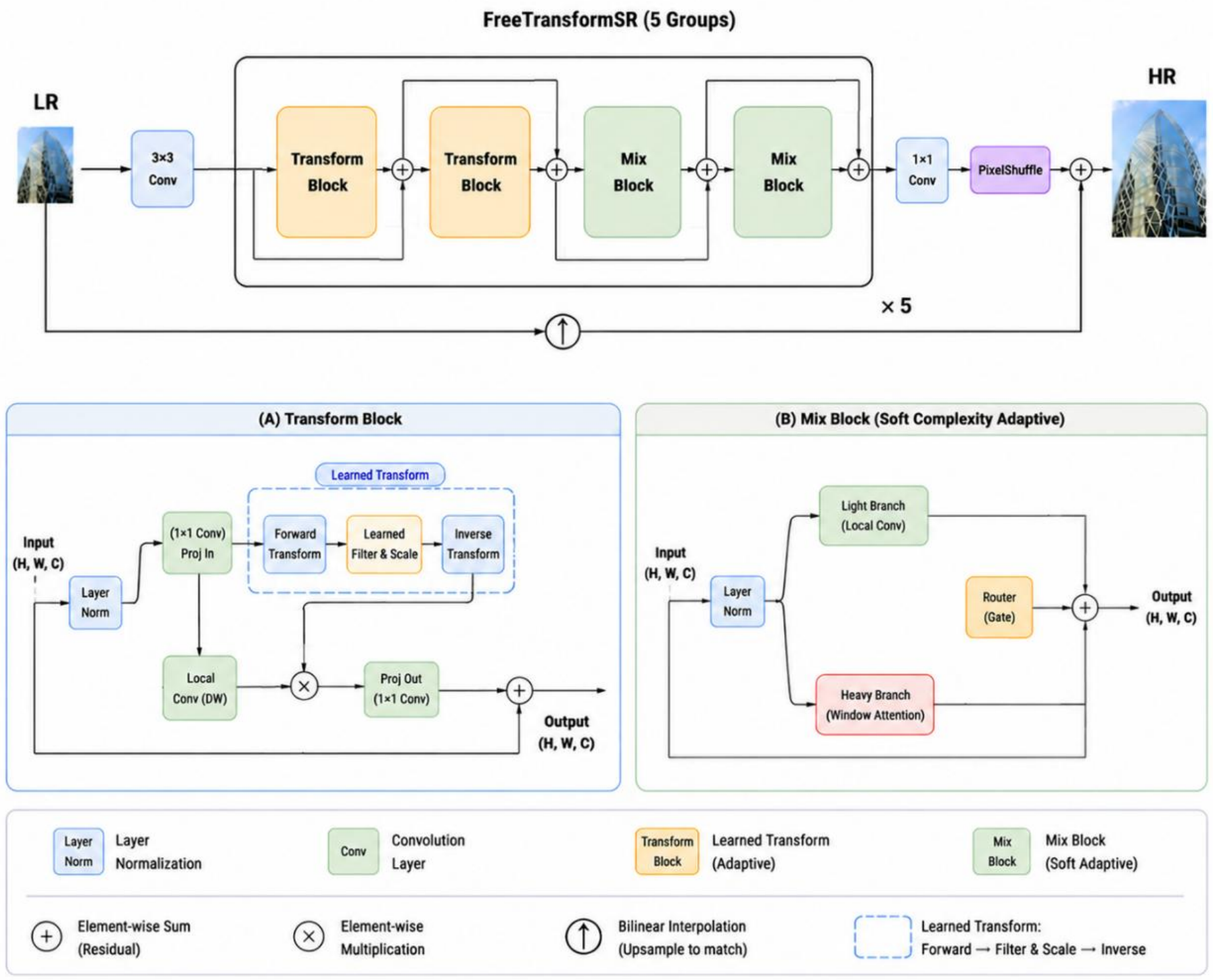

**Figure 1.** Overall architecture of the proposed FreeTransformSR network and the structures of its two core sub-modules.

The frequency filter bank and scale gate have distinct roles: the filter bank provides fixed per-channel frequency modulation templates, while the scale gate dynamically varies with input samples to determine the overall processing strength for the current sample. Together, they enable the network to flexibly adjust response intensity when facing inputs with varying degradation levels, achieving "sample-adaptive" feature modulation. The main contributions of this paper are summarized as follows:

- We propose a learnable channel-wise free low-rank transform that achieves adaptive feature modulation in a data-driven manner, distinguishing it from frequency-domain methods that rely on predefined bases.
- We design a soft complexity adaptive gating structure that adaptively fuses local convolution and window self-attention according to regional texture characteristics, dynamically selecting an appropriate fusion ratio.
- We introduce adaptive intensity modulation to dynamically adjust transform-domain response strength at the sample level, enhancing model adaptability to different degradation levels.
- Experimental results demonstrate that our method outperforms mainstream lightweight super-resolution schemes in terms of parameter count, FLOPs, and inference speed, while achieving competitive PSNR/SSIM performance.

## 2. Related Work

### *2.1. Lightweight Image Super-Resolution Networks*

In recent years, deep learning techniques have made significant progress in the field of image super-resolution. Early convolutional neural network-based methods continuously pushed performance boundaries by introducing residual learning, dense connections, and attention mechanisms. EDSR [4] removes batch normalization layers and stacks numerous residual blocks, significantly improving reconstruction accuracy while maintaining training stability. RDN [7] designs densely connected residual blocks to fully exploit multi-level and multi-scale features for high-quality reconstruction. RCAN [6] introduces channel attention mechanisms, enabling the network to adaptively recalibrate the importance of each channel, demonstrating clear advantages in handling complex textures. In addition, CARN [2] adopts a cascaded residual structure to achieve parameter-efficient utilization through cross-layer feature reuse. IMDN [3] proposes information multi-distillation blocks that decompose the feature extraction process into multiple fine-grained steps, maintaining competitive reconstruction performance under parameter constraints. RFDN [5] further introduces residual feature distillation on this basis, progressively refining feature representations to improve reconstruction quality. LatticeNet [23] designs a lattice block structure that enhances feature representation capability while maintaining lightweight characteristics. OSFFNet [24] proposes an omni-stage feature fusion strategy to fully leverage feature information across different levels. However, the aforementioned CNN methods are inherently limited by the local receptive field of convolution operations, struggling to adequately capture long-range pixel dependencies.

The powerful long-range modeling capability demonstrated by Transformers in high-level vision tasks has prompted researchers to introduce them into the super-resolution domain. SwinIR [8] first applies the Swin Transformer to image restoration, achieving global information interaction through window-based self-attention and shifted window strategies while controlling computational complexity. ELAN [12] further improves the efficiency of lightweight Transformers through cross-layer parameter sharing and grouped attention design. ESRT [9] adopts a two-stage feature extraction strategy to establish efficient communication between shallow and deep features. OmniSR [10] proposes an omnidirectional aggregation attention mechanism that simultaneously models long-range dependencies in both channel and spatial dimensions. ATD [11] introduces learnable token dictionaries to achieve token aggregation through adaptive dictionary lookup, breaking through the limitations of fixed windows. SRFormer [13] employs a permuted self-attention design to achieve global receptive field expansion with extremely low computational cost. HiT-SRF [14] designs a hierarchical attention distillation architecture to maximize the benefits of self-attention under limited parameter budgets. Furthermore, Restormer [25] introduces channel gating mechanisms in multi-head attention, dynamically adjusting the contribution weights of different channels based on feature response intensity, demonstrating excellent performance in image restoration tasks.

Despite the favorable FLOPs metrics of the above lightweight Transformer schemes, their practical inference efficiency remains severely constrained. The self-attention mechanism requires instantiating attention scores of size O(N^2) when computing similarity matrices, accompanied by memory-intensive operations such as tensor reshaping and window masking. This results in significantly higher actual latency and memory usage compared to CNN models with equivalent FLOPs. This issue is particularly prominent in super-resolution tasks, as such networks typically maintain feature map resolution without downsampling stages commonly found in vision Transformers.

### *2.2. Frequency-Domain, Transform-Domain, and Clustering Methods*

Frequency-domain analysis provides an alternative technical pathway for super-resolution reconstruction. Traditional methods typically employ discrete wavelet transform (DWT) or Fourier transform to decompose images into different frequency bands, separately processing low-frequency structures and high-frequency details. WavTrans [20] combines wavelet transform with cross-attention Transformers, enhancing texture details through frequency band separation in multi-contrast MRI super-resolution tasks. CiaoSR [21] constructs a continuous implicit attention network, incorporating frequency-domain prior information to guide detail generation during arbitrary-scale upsampling. ShuffleMixer [19] adopts large-kernel convolutions to simulate frequen-cy-domain decomposition effects in the spatial domain, achieving performance comparable to Transformers with a pure convolutional architecture.

However, the aforementioned frequency-domain methods all rely on predefined fixed basis functions (e.g., Haar wavelets, DCT, etc.), whose selection is independent of image content and lacks data-driven adaptive capability. To address this limitation, some studies have attempted to achieve data-adaptive token aggregation through clustering or hashing mechanisms. BOAT [26] proposes a bilateral local attention mechanism that, building upon spatial local attention, further employs feature-space clustering to aggregate similar patches into clusters and compute attention within each cluster, effectively capturing cross-window long-range patch correlations. TCFormer [27] merges tokens from different positions with flexible shapes and sizes through progressive clustering, enabling the model to adaptively adjust token shapes according to semantic concepts. Clustered Attention [28] groups queries into clusters and computes attention only on cluster centers, achieving sparse attention approximation with linear complexity. NLSA [17] partitions the input space into hash buckets using spherical locality-sensitive hashing, with each query performing non-local attention computation only within its assigned bucket, reducing computational complexity from quadratic to asymptotically linear. SPIN [16] achieves token aggregation through soft K-means clustering [22], using cluster centers as proxies for attention computation. The above clustering methods reduce the computational cost of self-attention through data-driven grouping strategies to varying degrees, but they also face issues such as increased inference latency due to clustering iterations and loss of detail information from coarse-grained cluster center representations.

Dynamic convolution [18] explores adaptive feature transformation from another perspective, dynamically combining multiple convolution kernels through input-dependent attention weights. However, dynamic convolution can only achieve input-adaptive adjustment of kernel weights and cannot alter the inherent frequency response characteristics of basis functions.

The channel-wise learnable transform proposed in this paper is fundamentally different from the aforementioned methods: it independently learns 2D transformation matrices for each feature channel and parameterizes them in a low-rank residual form, thereby achieving truly data-driven basis function evolution with extremely low parameter overhead, breaking through the representation bottleneck of fixed basis functions.

### *2.3. Dynamic Networks and Adaptive Mechanisms*

Dynamic networks establish a flexible balance between efficiency and accuracy by adaptively adjusting network structures or parameters according to input content. The aforementioned channel attention mechanism [6] can be viewed as an early form of dynamic response, achieving channel-wise feature recalibration through global pooling and gated convolution. In addition, Han et al. theoretically established the connection between local attention and dynamic depthwise separable convolution, demonstrating that dynamic convolution can simulate the behavioral patterns of attention mechanisms under certain conditions.

The adaptive intensity modulation mechanism (scale gate) proposed in this paper provides a complementary design to the above dynamic mechanisms: it extracts modulation coefficients from the overall features of each input sample through global average pooling and a lightweight gating network, performing global scaling on the response magnitude of transform-domain outputs. The modulation coefficients are automatically guided by the loss function during network training without explicit rule constraints—the network autonomously determines which inputs should be amplified or suppressed based on the statistical characteristics of the training data, resulting in data-driven implicit adaptive behavior. Unlike feature reweighting or path selection, this mechanism directly operates on the overall response intensity of the transform domain while introducing only a minimal number of additional parameters.

Furthermore, recent advancements in dynamic networks have also explored Mixture-of-Experts (MoE) architectures to improve inference efficiency. While these methods share a similar philosophy of dynamic computation, our soft complexity adaptive mechanism functions as a continuous mixing ratio rather than discrete routing, offering an alternative perspective on balancing efficiency and reconstruction quality.

In summary, existing lightweight super-resolution methods exhibit a significant imbalance among efficiency, adaptability, and generalization capability. FreeTransformSR innovates synergistically from three dimensions—learnable transform, adaptive fusion, and robust training—providing a more balanced solution for lightweight super-resolution tasks.

## 3. Methods

In this section, we detail the proposed FreeTransformSR network. Section 3.1 presents the overall architecture, Section 3.2 elaborates on the design and implementation of the core learnable transform module, Section 3.3 introduces the soft complexity adaptive module, and Section 3.4 describes the training loss function.

### *3.1. Overall Network Architecture*

The overall architecture of FreeTransformSR is illustrated in Figure 1, which mainly consists of three modules: shallow feature extraction, deep feature extraction, and image reconstruction.

Given a low-resolution input image $I_{LR} \in \mathbb{R}^{H\times W\times 3}$, we first apply a 3×3 convolution for shallow feature extraction, mapping the input from RGB space to a high-dimensional feature space:

$$F_0 = Conv_{3\times 3}(I_{LR}), F_0 \in \mathbb{R}^{H\times W\times C} \quad (1)$$

Where C is the number of feature channels (set to 60 in this work), and H and W denote the height and width of the input image, respectively.

The shallow features $F_0$ are then sequentially passed through N residual groups (RGs) for deep feature extraction. Each residual group consists of several basic blocks, which are categorized into two types: Learned Transform Blocks (LTB) and Soft Complexity Adaptive Blocks (SCAB). The two types of blocks are arranged in a specific order within each residual group: each group contains 4 basic blocks, with the first 2 being LTBs and the remaining 2 being SCABs. At the end of each residual group, a 1×1 convolution and an Efficient Channel Attention (ECA) [29] module are applied for feature aggregation and recalibration. The computation process of the i-th residual group can be formulated as:

$$F_i = F_{i-1} + ECA\left(Conv_{1\times1}(\mathcal{B}_i(F_{i-1}))\right), i = 1,2,\ldots,N \tag{2}$$

Where $\mathcal{B}_i(\cdot)$ denotes the sequential operation of basic blocks within the i-th residual group, and the residual connection is used to stabilize the training process.

After all residual groups, the deep features are further refined by a 1×1 convolution:

$$F_{refine} = Conv_{1\times1}(F_N) \tag{3}$$

Finally, the image reconstruction module consists of a 3×3 convolution followed by a PixelShuffle [30] operation, which upsamples the refined features to the target resolution:

$$I_{SR} = PixelShuffle\left(Conv_{3\times3}(F_{refine})\right) + Bilinear(I_{LR}) \tag{4}$$

Where $Bilinear(\cdot)$ denotes bilinear upsampling, serving as a global residual connection to provide low-frequency content prior for the reconstruction process.

### *3.2. Learnable Transform Block*

We design a channel-wise learnable transform that independently learns a pair of 2D transformation matrices for each feature channel, parameterized in the form of "identity mapping + low-rank residual". A local feature modulation branch is also introduced to enhance high-frequency detail recovery via depthwise convolution.

#### 3.2.1. Free Low-Rank Transform

Existing frequency-domain methods typically rely on predefined fixed basis functions (e.g., discrete wavelet transform, discrete cosine transform, etc.) for frequency decomposition. These basis functions are determined before training and are independent of image content, lacking data-driven adaptive capability. Moreover, their forms are constrained by mathematical constructions (e.g., orthogonality of Haar wavelets, symmetry of DCT), making it difficult to optimize them for specific tasks. To address these limitations, we design a channel-wise learnable transform that independently learns a pair of 2D transformation matrices for each feature channel, parameterized as "identity mapping + low-rank residual", while introducing a local feature modulation branch to enhance high-frequency detail recovery.

Let the input feature be $X \in \mathbb{R}^{B\times C\times h\times w}$, We first apply a 1×1convolution to expand the channel dimension by a factor of 2, and then split it along the channel dimension into $X_1$ and $X_2$[31]. $X_1$ is fed into the transform domain for global modulation, while $X_2$ is sent to the local convolution branch for high-frequency detail extraction.

We partition $X_1$ into windows of size $p \times p$ (with $p = 8$ in this work). The choice of a small window size (p=8) for the learnable transform is motivated by the need for fine-grained frequency modulation: smaller windows enable the transform to capture local texture and high-frequency details more precisely, while keeping the parameter count low, consistent with our lightweight design. For each window $X_{win} \in \mathbb{R}^{B\times C\times p\times p}$, we learn a pair of transformation matrices $T_h^{(c)} \in \mathbb{R}^{p\times p}$ (row transform) and $T_w^{(c)} \in \mathbb{R}^{p\times p}$ (column transform) for each channel $c$.

To reduce parameter redundancy and improve training efficiency, we design the transformation matrices in a low-rank form:

$$T_h^{(c)} = I + \Delta_h^{(c)}, T_w^{(c)} = I + \Delta_w^{(c)} \tag{5}$$

Where $I \in \mathbb{R}^{p\times p}$ is the identity matrix, and $\Delta_h^{(c)}, \Delta_w^{(c)} \in \mathbb{R}^{p\times p}$ are learnable low-rank residual matrices. Specifically, $\Delta_h^{(c)}$ and $\Delta_w^{(c)}$ are each decomposed into the product of two low-rank factors:

$$\Delta_h^{(c)} = A_h^{(c)} B_h^{(c)}, \Delta_w^{(c)} = A_w^{(c)} B_w^{(c)} \tag{6}$$

Where $A_h^{(c)} \in \mathbb{R}^{p\times r}, B_h^{(c)} \in \mathbb{R}^{r\times p}, A_w^{(c)} \in \mathbb{R}^{p\times r}, B_w^{(c)} \in \mathbb{R}^{r\times p}$, and ris the rank (set to $r = 1$ in this work). All low-rank factors are trainable parameters initialized to zero, so both $T_h^{(c)}$ and $T_w^{(c)}$ start as identity matrices, and the module begins learning from the identity mapping.

The forward transform maps each window block from the spatial domain to the transform domain:

$$Y = T_h X_{win} T_w^T \tag{7}$$

Where $Y \in \mathbb{R}^{B\times C\times p\times p}$ is the transform-domain feature. In the transform domain, we apply per-channel and per-frequency modulation using a learnable coefficient matrix $W \in \mathbb{R}^{C\times p\times p}$ (named coeff in the code):

$$\tilde{Y} = Y \odot \boldsymbol{W} \odot \boldsymbol{S} \tag{8}$$

Where $\odot$ denotes element-wise multiplication. $\boldsymbol{W}$ is a learnable coefficient fixed in the network parameters, providing a base amplitude bias for each channel and frequency position.

Additionally, we introduce an adaptive intensity modulation module that extracts global responses from the input features and generates a sample-level scaling coefficient $\boldsymbol{S}$:

$$S = 0.8 + 0.6 \cdot \sigma(Conv_{1\times1}(AvgPool(X))) \tag{9}$$

where $\sigma(\cdot)$ is the Sigmoid activation function and AvgPool$(\cdot)$ denotes global average pooling. This module first aggregates spatial information across channels via global average pooling, then uses a $1 \times 1$ convolution to linearly weight and aggregate the $C$ channel responses into a single channel, yielding a unified scalar $\boldsymbol{S} \in \mathbb{R}^{B\times1\times1\times1}$. Although this scalar is derived from information across all channels, it is applied uniformly to all channels to control the overall processing intensity of the current sample. $\boldsymbol{W}$ and $\boldsymbol{S}$ have distinct roles: $\boldsymbol{W}$ provides a fixed per-channel frequency modulation template, while $\boldsymbol{S}$ varies dynamically with the input sample to determine the overall amplification or suppression of transform-domain responses. Together, they enable the network to flexibly adjust response intensity when facing inputs of varying quality.

Finally, the features are recovered from the transform domain to the spatial domain via transposed back-mapping:

$$\hat{X}_{win} = T_h^T \tilde{Y} T_w \tag{10}$$

This operation involves only matrix multiplication, which reduces the computational complexity to $O(p^2)$and avoids imposing extra constraints on the transformation matrices. Experiments demonstrate that this strategy is highly effective in super-resolution tasks: it provides a larger parameter space for the network, leading to better reconstruction quality while significantly reducing computational costs. After reorganizing all window blocks $\hat{X}_{win}$back to their original spatial positions, we obtain the complete feature map $\hat{X}_1$.

3.2.2. Local Feature Modulation

After transform-domain modulation and back-mapping, we introduce a local feature modulation branch to enhance high-frequency detail recovery. Specifically, we apply depthwise separable convolution to $\boldsymbol{X}_2$ (which has not undergone transform-domain processing) to extract local spatial features:

$$X_{local} = Conv_{3\times3}^{depthwise}(X_2) \tag{11}$$

Subsequently, the local features are multiplied element-wise with the transform-domain recovered features $\hat{\boldsymbol{X}}_1$, enabling local details to modulate global features:

$$X_{mod} = \hat{X}_1 \odot X_{local} \tag{12}$$

Finally, a $1 \times 1$ convolution is applied for feature fusion. The core idea of this design is that the learnable transform excels at capturing low-frequency structures and global correlations, while depthwise convolution is adept at extracting high-frequency textures and local edges. Through element-wise multiplication, the local detail features perform gated enhancement on the global transform features, enabling the network to maintain global consistency while preserving high-frequency details. This design draws inspiration from the frequency feature aggregation concept in EFATSR [31], using an explicit local convolution branch to supplement high-frequency information that may be lost in transform-domain processing.

In summary, the learnable transform block completes feature enhancement through the pipeline of "forward transform → transform-domain modulation → transposed back-mapping → local feature modulation", and its complete forward process is uniformly described by Equations (5) through (12).

### *3.3. Soft Complexity Adaptive Module*

In lightweight super-resolution network design, a common practice is to increase the number of channels to obtain larger feature capacity. However, our experimental observations indicate that when relying solely on stacked window self-attention modules, the performance gain from simply increasing channel numbers rapidly diminishes—the marginal return of additional parameters does not justify the computational cost. Motivated by this, we explore improving performance by enriching feature representation pathways rather than merely widening channels.

Specifically, we design a soft complexity adaptive module that, while keeping the total parameter count essentially unchanged, reallocates some parameters originally used for channel stacking to a lightweight convolution branch, and dynamically fuses it with the window self-attention branch through a gating mechanism. This design allows two operations with different inductive biases to work synergistically: local convolution excels at preserving edge sharpness and structural consistency, while window self-attention excels at capturing long-range texture dependencies. They are softly weighted and fused through a lightweight gating network based on input features, enabling the model to adaptively select appropriate processing approaches in different regions.

The local convolution branch $L(\cdot)$ consists of layer normalization, $3 \times 3$ depthwise separable convolution, and $1 \times 1$ pointwise convolution, excelling at extracting local edge and texture information:

$$L(X) = Conv_{1\times1}\left(GELU\left(Conv_{3\times3}^{depthwise}\left(LayerNorm(X)\right)\right)\right) + X \tag{13}$$

The window self-attention branch $\mathcal{H}(\cdot)$ adopts the standard shifted window self-attention mechanism [8], with a window size of $16 \times 16$ and 3 heads, effectively capturing long-range dependencies within windows:

$$A(X) = WindowAttention\big(LayerNorm(X)\big) + X \tag{14}$$

The self-attention window (16×16) is twice the size of the learnable transform (p=8), forming a hierarchical spatial scale design. The 16×16 attention window covers a 2×2 region of transform windows, enabling the attention mechanism to model long-range spatial dependencies across multiple transform patches, while the transform performs fine-grained frequency modulation. The two modules complement each other at different spatial scales.

The gating network $\mathcal{R}(\cdot)$ generates spatially varying fusion coefficients based on input features. It first extracts features through depthwise convolution and two $1 \times 1$ convolutions, then maps them to the $[0.1, 0.9]$ range via Sigmoid activation:

$$\mathcal{R}(X) = 0.1 + 0.8 \cdot \sigma(Conv_{1\times1}\left(GELU\left(Conv_{1\times1}\left(Conv_{3\times3}^{depthwise}(X)\right)\right)\right)) \tag{15}$$

where $\sigma(\cdot)$ is the Sigmoid activation function. We constrain the gating values to the [0.1, 0.9] range to prevent either branch from being completely deactivated due to gate saturation during training, ensuring that gradients can continuously flow through both branches. Specifically, the constant 0.1 in Eq. (19) serves as the lower bound ($\gamma_{min}$), guaranteeing that the local convolution branch always receives at least 10% of its output and thus avoiding gradient starvation. The constant 0.8 represents the dynamic range ($\gamma_{max} - \gamma_{min} = 0.9 - 0.1 = 0.8$), which provides sufficient adaptability for the network to learn meaningful fusion patterns while keeping the gate values bounded away from the extremes 0 and 1 to promote training stability. Moreover, the symmetric choice around 0.5 reflects our design principle that neither branch should be a priori favored—the network autonomously learns the optimal fusion balance from data through end-to-end optimization. Finally, the output of the soft complexity adaptive module is:

$$X_{out} = L(X) + \mathcal{R}(X) \odot (A(X) - L(X)) \tag{16}$$

This gating mechanism computes a continuous fusion coefficient based on input features, enabling the model to adaptively select an appropriate fusion ratio according to regional image characteristics—when the gating value approaches 0, the output favors the local convolution branch; when it approaches 1, the output favors the window self-attention branch; intermediate values achieve a soft blend of the two. Unlike hard selection, this soft fusion approach allows gradients to flow through both branches simultaneously, enabling them to mutually promote each other during training.

It is worth noting that the gating network itself does not contain an explicit "texture complexity" prior; its behavior is entirely driven by the training data—the network learns through the guidance of the loss function to determine which regions should favor which branch.

### *3.4. Loss Function*

To simultaneously constrain the quality of reconstructed images in both the spatial and frequency domains, we adopt a weighted combination of L1 pixel loss and Fourier space loss as the total loss function.

**L1 Pixel Loss** measures the per-pixel absolute error between the reconstructed image $I_{SR}$ and the high-resolution ground truth $I_{HR}$:

$$\mathcal{L}_{L1} = \frac{1}{N}\sum_{i=1}^{N}\left\|I_{SR}^{(i)} - I_{HR}^{(i)}\right\|_1 \tag{17}$$

where $N$ is the total number of pixels. Compared to L2 loss, L1 loss is less sensitive to outliers and produces sharper edges, making it widely used as a basic supervision signal for image restoration tasks [32].

**Fourier Space Loss** constrains reconstruction quality in the frequency domain. Let $\mathcal{F}(\cdot)$ denote the 2D Fast Fourier Transform (FFT). The Fourier space loss is defined as the mean squared error between the reconstructed image and the ground truth in the frequency domain [33]:

$$\mathcal{L}_{FFT} = \frac{1}{N}\sum_{i=1}^{N}\left\|\mathcal{F}(I_{SR}^{(i)}) - \mathcal{F}(I_{HR}^{(i)})\right\|_2^2 \quad (18)$$

The Fourier space loss directly supervises the recovery of frequency-domain information, effectively guiding the network to focus on the reconstruction of high-frequency components. This loss complements the L1 pixel loss—L1 loss constrains per-pixel accuracy in the spatial domain, while Fourier loss guides the consistency of frequency-domain distribution, together promoting convergence in both domains [33].

The total loss function is the weighted sum of these two losses:

$$\mathcal{L}_{total} = \mathcal{L}_{L1} + \lambda \cdot \mathcal{L}_{FFT} \quad (19)$$

where $\lambda = 0.1$ is the weight coefficient for the Fourier loss. This weighting ensures that the model primarily optimizes for spatial-domain reconstruction while using frequency-domain supervision as an auxiliary guide, enhancing high-frequency detail recovery without interfering with the convergence of the main task.

## 4. Experiments

In this section, we conduct comprehensive experiments to evaluate FreeTransformSR. We first introduce the experimental setup, then compare FreeTransformSR with state-of-the-art lightweight super-resolution methods through quantitative and qualitative analyses. Ablation studies are subsequently performed to validate the effectiveness of each core module, and finally, we report model inference speed. All experiments are conducted on an NVIDIA RTX 5090 GPU.

### *4.1. Experimental Settings*

Datasets. We adopt DIV2K [34] as the training set, which contains 800 high-resolution training images. The test sets include Set5 [35], Set14 [36], BSD100 [37], Urban100 [38], and Manga109 [39], covering diverse scenes and texture complexities.

**Training Details.** The proposed FreeTransformSR consists of $N = 5$ residual groups, each containing 4 basic blocks, and the number of feature channels is set to 60. During training, we randomly crop $128 \times 128$patches from HR images, with corresponding LR regions cropped according to the scaling factor. The AdamW optimizer is used with an initial learning rate of $1 \times 10^{-3}$, decayed to $1 \times 10^{-7}$via cosine annealing over 400K iterations. Data augmentation includes random horizontal/vertical flips and $90°/270°$rotations. The loss function is a weighted combination of L1 loss and Fourier space loss, with the Fourier loss weight set to 0.1.

Evaluation Metrics. PSNR and SSIM [40] are used as quantitative evaluation metrics, computed on the Y channel in the YCbCr color space. Model complexity is measured by parameter count (Params) and FLOPs, where FLOPs are calculated when the model upsamples an LR image to 1280 × 720 resolution.

### *4.2. Comparison with State-of-the-Art Methods*

We compare FreeTransformSR with current state-of-the-art lightweight super-resolution methods, including CARN [2], SwinIR-light [8], ESRT [9], SRFormer-light [13], CAMixerSR [41], and MambaIRv2-light [42]. The experimental results of all compared methods are cited from their original papers.

Quantitative Results. Table 1 reports the PSNR/SSIM results of FreeTransformSR at ×2, ×3, and ×4 scales, respectively. FreeTransformSR achieves the smallest parameter count and FLOPs among all compared methods while attaining optimal performance on most datasets.

At ×2 scale, FreeTransformSR (634K, 206.0G) achieves 39.51 dB on Manga109, surpassing MambaIRv2-light (39.35 dB) with only 74.4% of its parameters. Compared to SRFormer-light (853K, 236.2G), FreeTransformSR improves Manga109 PSNR by 0.23 dB with 32.5% fewer parameters. At ×3 and ×4 scales, FreeTransformSR consistently demonstrates leading performance. On Urban100 ×4, FreeTransformSR outperforms SwinIR-light by 0.53 dB with 54.0G FLOPs and 653K parameters, whereas SwinIR-light has 63.6G FLOPs and 930K parameters. These results verify that the learnable transform achieves superior performance with significantly smaller model sizes.

**Table 1.** Quantitative comparison with state-of-the-art methods (PSNR/SSIM). Best and second-best results are highlighted in red and blue, respectively.

| Method | Scale | Params | FLOPs | Set5 | Set14 | BSD100 | Urban100 | Manga109 |
|---|---|---|---|---|---|---|---|---|
| | | | | PSNR/SSIM | PSNR/SSIM | PSNR/SSIM | PSNR/SSIM | PSNR/SSIM |
| CARN [2] | ×2 | 1,592K | 222.8G | 37.76/0.9590 | 33.52/0.9166 | 32.09/0.8978 | 31.92/0.9256 | 38.36/0.9765 |
| SwinIR-light [8] | ×2 | 910K | 244.2G | 38.14/0.9611 | 33.86/0.9206 | 32.31/0.9012 | 32.76/0.9340 | 39.12/0.9783 |
| ESRT [9] | ×2 | 677K | 191.4G | 38.03/0.9600 | 33.75/0.9184 | 32.25/0.9001 | 32.58/0.9318 | 39.12/0.9774 |
| SRFormer-light [13] | ×2 | 853K | 236.2G | 38.23/0.9613 | 33.94/0.9209 | 32.36/0.9019 | 32.91/0.9353 | 39.28/0.9785 |
| CAMixerSR [41] | ×2 | 746K | 204.7G | 38.23/0.9613 | 34.00/0.9214 | 32.34/0.9016 | 32.95/0.9348 | 39.32/0.9781 |
| MambaIRv2-light [42] | ×2 | 774K | 286.3G | 38.26/0.9615 | 34.09/0.9221 | 32.36/0.9019 | 33.26/0.9378 | 39.35/0.9785 |
| **FreeTransformSR (our)** | ×2 | **634K** | **206.0G** | **38.33/0.9618** | **34.10/0.9225** | **32.41/0.9027** | **33.26/0.9384** | **39.51/0.9785** |
| CARN [2] | ×3 | 1,592K | 118.8G | 34.29/0.9255 | 30.29/0.8407 | 29.06/0.8034 | 28.06/0.8493 | 33.50/0.9440 |
| SwinIR-light [8] | ×3 | 918K | 111.2G | 34.62/0.9289 | 30.54/0.8463 | 29.20/0.8082 | 28.66/0.8624 | 33.98/0.9478 |
| ESRT [9] | ×3 | 770K | 96.4G | 34.42/0.9268 | 30.43/0.8433 | 29.15/0.8063 | 28.46/0.8574 | 33.95/0.9455 |
| SRFormer-light [13] | ×3 | 861K | 105.4G | 34.67/0.9296 | 30.57/0.8469 | 29.26/0.8099 | 28.81/0.8655 | 34.19/0.9489 |
| CAMixerSR [41] | ×3 | 754K | 89.7G | 34.66/0.9296 | 30.62/0.8471 | 29.25/0.8092 | 28.81/0.8644 | 34.34/0.9491 |
| MambaIRv2-light [42] | ×3 | 781K | 126.7G | 34.71/0.9298 | 30.68/0.8483 | 29.26/0.8098 | 29.01/0.8689 | 34.41/0.9497 |
| **FreeTransformSR (our)** | ×3 | **642K** | **92.2G** | **34.84/0.9311** | **30.74/0.8495** | **29.34/0.8119** | **29.18/0.8716** | **34.71/0.9511** |
| CARN [2] | ×4 | 1,592K | 90.9G | 32.13/0.8937 | 28.60/0.7806 | 27.58/0.7349 | 26.07/0.7837 | 30.47/0.9084 |
| SwinIR-light [8] | ×4 | 930K | 63.6G | 32.44/0.8976 | 28.77/0.7858 | 27.69/0.7406 | 26.47/0.7980 | 30.92/0.9151 |
| ESRT [9] | ×4 | 751K | 67.7G | 32.19/0.8947 | 28.69/0.7833 | 27.69/0.7379 | 26.39/0.7962 | 30.75/0.9100 |
| SRFormer-light [13] | ×4 | 873K | 62.8G | 32.51/0.8988 | 28.82/0.7872 | 27.73/0.7422 | 26.67/0.8032 | 31.17/0.9165 |
| CAMixerSR [41] | ×4 | 765K | 54.5G | 32.51/0.8988 | 28.82/0.7870 | 27.72/0.7416 | 26.63/0.8012 | 31.18/0.9166 |
| MambaIRv2-light [42] | ×4 | 790K | 75.6G | 32.51/0.8992 | 28.84/0.7878 | 27.75/0.7426 | 26.82/0.8079 | 31.24/0.9182 |
| **FreeTransformSR (our)** | **×4** | **653K** | **54.0G** | **32.65/0.9010** | **28.98/0.7906** | **27.84/0.7449** | **27.00/0.8112** | **31.61/0.9205** |

**Visual Comparison. Figure 2** presents the visual results of different methods on Urban100, Set14, and Manga109. The compared methods include CARN [2], ESRT [9], SRFormer-light [13], CAMixerSR [41], and MambaIRv2-light [42]. FreeTransformSR reconstructs images with sharper edges, more realistic textures, and fewer artifacts across all datasets. In regions with repetitive textures (e.g., grid patterns in Urban100) or fine details (e.g., the parrot feathers in Set14 and text strokes in Manga109), FreeTransformSR more accurately recovers structural details, while competing methods still exhibit blur or distortion. This benefits from the adaptive modulation of global frequency information by the learnable transform and the high-frequency detail supplementation by the local feature modulation branch.

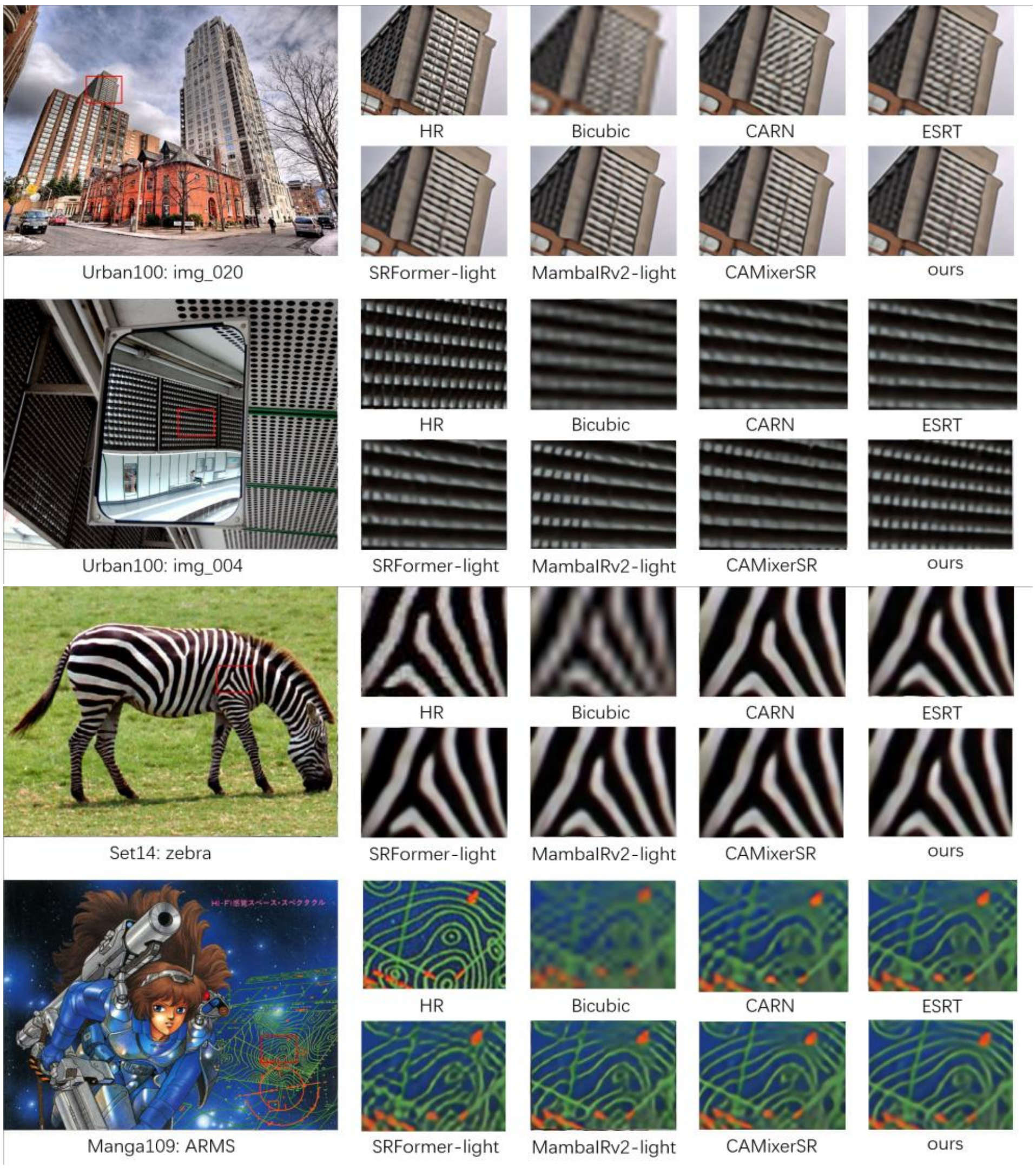

**Figure 2.** Visual comparison of FreeTransformSR and existing methods on ×4 super-resolution tasks.

### *4.3. Ablation Study*

To validate the effectiveness of each core module, we train all ablation variants on DIV2K for 400K iterations and evaluate ×2 performance on Set5, Set14, and Urban100. All variants adopt the same training configuration to ensure fair comparison.

We design ten variants:

- w/o both: removes both local feature modulation and adaptive intensity modulation.
- Single-Attention: removes only the local feature modulation branch (i.e., removes the element-wise multiplication between $x_1$ and the local convolution output), relying solely on the non-local attention mechanism for feature aggregation, while retaining adaptive intensity modulation.
- Single-Conv: removes the non-local attention branch entirely, relying solely on local convolutions.
- w/o scale gate: removes adaptive intensity modulation (fixing the scaling coefficient to 1.0), while retaining local feature modulation.
- Balanced: based on w/o Local, increases the channel number from 60 to 63, raising the parameter count to 629.29K, comparable to Full (633.83K), to rule out the possibility that performance differences are merely due to insufficient parameters.
- Single Rank: all channels share a single transformation matrix (i.e., a dim=1 version of the transform with rank r=1), removing channel-independent transforms to verify the necessity of channel-independent learning.

- Full Rank: uses the full transformation matrix (without low-rank constraint, i.e., without decomposing Δ into low-rank factors), but all channels still share the same transformation matrix (578.15K parameters), to verify whether the low-rank constraint limits model capacity.
- Fourier: replaces the learnable transform with a fixed Discrete Cosine Transform (DCT) as a baseline for fixed basis transform comparison (624.23K parameters, 201.66G FLOPs).
- All-Hybrid: replaces all LTB blocks with SCAB blocks (i.e., four SCAB blocks per residual group), yielding 665.33K parameters and 294.51G FLOPs, to evaluate the effect of purely attention-heavy design.
- All-Transform: replaces all SCAB blocks with LTB blocks (i.e., four LTB blocks per residual group), yielding 602K parameters and 117.57G FLOPs, to evaluate the effect of purely convolution-heavy design.
- Full: includes all components with 60 channels, employing channel-independent + low-rank (r=1) design.

Table 2 reports the ablation results.

**Table 2.** Ablation study on core modules (×2).

| Variant | Description | Params | FLOPs | Set5 | Set14 | Urban100 |
|---|---|---|---|---|---|---|
| | | | | PSNR/SSIM | PSNR/SSIM | PSNR/SSIM |
| w/o both | Local ✗, SG ✗ | 574.11K | 193.01G | 38.31/0.9618 | 34.03/0.9216 | 33.19/0.9375 |
| Single-Attn | Local ✗ (Attn Only), SG ✓ | 574.72K | 193.15G | 38.31/0.9618 | 34.06/0.9216 | 33.25/0.9377 |
| Single-Conv | Local ✗ (Conv Only), SG ✓ | 470.92K | 96.42G | 38.08/0.9609 | 33.76/0.9190 | 32.46/0.9309 |
| w/o scale gate | Local ✓, SG ✗ | 633.22K | 205.90G | 38.32/0.9618 | 33.98/0.9216 | 33.26/0.9377 |
| Balanced | dim: 60→63 (w/o Local) | 629.29K | 208.79G | 38.32/0.9617 | 34.04/0.9220 | 33.24/0.9376 |
| Single Rank | shared transform, r=1 | 577.19K | 206.04G | 38.31/0.9617 | 34.04/0.9215 | 33.24/0.9374 |
| Full Rank | shared transform, no low-rank | 578.15K | 206.04G | 38.31/0.9617 | 34.03/0.9213 | 33.26/0.9382 |
| Fourier | fixed DCT transform | 624.23K | 201.66G | 38.32/0.9618 | 34.07/0.9222 | 33.14/0.9371 |
| All-Hybrid | 4× SCAB blocks | 665.33K | 294.51G | 38.33/0.9618 | 34.10/0.9220 | 33.30/0.9382 |
| All-Transform | 4× LTB blocks | 602.00K | 117.57G | 38.13/0.9610 | 33.77/0.9187 | 32.57/0.9315 |
| Full | Local ✓, SG ✓, r=1 | 633.83K | 206.04G | 38.33/0.9618 | 34.10/0.9225 | 33.26/0.9384 |

**Effect of Local Feature Modulation and Inductive Bias.** Comparing Single-Attention and Full, adding the local feature modulation branch yields a slight performance gain (0.04 dB on Set14 and 0.01 dB on Urban100). This branch introduces approximately 59K additional parameters (633.83K vs. 574.72K) and 12.9G FLOPs. Furthermore, comparing Single-Conv and Single-Attention reveals that the pure convolution variant is highly lightweight, capable of achieving baseline reconstruction performance, and can effectively synergize with the attention mechanism for further enhancement. The Full model further surpasses Single-Attention, validating that our gating mechanism successfully synergizes the local detail extraction of convolution with the global context modeling of attention, yielding the best overall performance.

**Effect of Adaptive Intensity Modulation.** Comparing w/o scale gate and Full, the scale gate improves Set14 by 0.12 dB with negligible parameter overhead (633.83K vs 633.22K), demonstrating its high parameter efficiency as a content-adaptive scaling mechanism.

**Parameter Count Balancing Verification.** To rule out the confounding effect of parameter count differences, we conducted the following comparative analysis. Comparing Balanced with w/o Local, merely increasing the channel width (60→63) to match the parameter count adds approximately 54.57K parameters and 15.6G FLOPs, yet yields no substantial performance gain, indicating that simply widening channels on this baseline offers diminishing returns. Further comparing Balanced with Full, both models have nearly identical parameter counts (629.29K vs. 633.83K), but Full outperforms Balanced by 0.06 dB on Set14 and 0.02 dB on Urban100 while reducing FLOPs by approximately 2.75G.

These results confirm that the performance advantage of Full stems from its architectural design rather than an increase in model size.

**Effectiveness of Low-Rank Constraint.** Comparing Single Rank (shared transform, r=1) and Full Rank (shared transform, no low-rank constraint), the two achieve nearly identical performance across all three datasets—Full Rank leads by 0.02 dB on Urban100 (33.26 vs 33.24), while Single Rank leads by 0.01 dB on Set14 (34.04 vs 34.03). This demonstrates that the low-rank constraint does not cause perceptible performance loss, yet reduces parameters from 578.15K to 577.19K, validating low-rank decomposition as an effective parameter efficiency strategy—achieving feature representation capability comparable to the full transformation matrix with minimal parameter cost. Further comparing Single Rank (shared transform, r=1) and Full (channel-independent, r=1), the latter improves Set14 by 0.06 dB, indicating that channel-independent learning yields more significant gains than shared transforms, which justifies Full's adoption of channel-independent design.

**Comparison with Fixed Basis Transform.** Comparing Fourier (fixed DCT) and Full, the fixed DCT achieves 33.14 dB on Urban100, validating the effectiveness of frequency-domain processing for super-resolution. Full leads by 0.12 dB on Urban100 and 0.03 dB on Set14, demonstrating that the data-driven learnable transform better adapts to the feature modulation demands of super-resolution tasks compared to fixed basis transforms.

**Combined Effect of Modules.** Comparing w/o both and Full, the two modules together contribute 0.07 dB improvement on Set14 and 0.07 dB on Urban100, with ablation variants showing a progressive upward trend in performance.

**Block Configuration Ratio.** To justify the $2 + 2$design, we evaluated two extreme configurations: All-Hybrid ($4 \times$ SCAB) and All-Transform ($4 \times$ LTB). As reported in Table 2, All-Transform achieves an extremely low computational cost (117.57G FLOPs, a 43% reduction compared to Full), with only a slight compromise in reconstruction performance (Set14: -0.33 dB; Urban100: -0.69 dB). Conversely, All-Hybrid brings a marginal performance improvement (Set14: equal; Urban100: +0.04 dB), but at the expense of a substantially higher computational overhead (+43% FLOPs). By contrast, the $2 + 2$configuration effectively combines the advantages of both, achieving a favorable balance between reconstruction accuracy and computational efficiency.

### 4.4. *Inference Latency Comparison*

We selected three lightweight models (CAMixerSR [41], MambaIRv2-light [42], and SRFormer-light [13]) with comparable performance to ours for latency comparison. We measured their inference latency on an NVIDIA RTX 5090 GPU, with LR images as input at the corresponding scaling factor and HR images at $1280 \times 720$resolution as output. The reported values are averaged over 100 inference runs (unit: ms/img). The detailed results are reported in Table 3.

**Table 3.** Throughput comparison (ms/img).

| Method | Throughput (×2) | Throughput (×4) |
| --- | --- | --- |
| SRFormer-light [13] | 139.14 ms/img | 37.58 ms/img |
| CAMixerSR [41] | 99.40 ms/img | 43.71 ms/img |
| MambaIRv2-light [42] | 202.84 ms/img | 54.41 ms/img |
| FreeTransformSR (Ours) | 93.81 ms/img | 36.60 ms/img |

As shown in Table 3, FreeTransformSR achieves the lowest inference latency among the evaluated models at both scales (93.81 ms/img at ×2 and 36.60 ms/img at ×4). Compared to CAMixerSR, FreeTransformSR is 7.11 ms faster at ×4. Compared to MambaIRv2-light, FreeTransformSR is 2.16× faster at ×2.

## 5. Discussion

This paper compares FreeTransformSR with fixed-basis frequency methods and purely attention/convolution-based networks. The results demonstrate that our data-driven low-rank transform and adaptive gating mechanism more effectively recover high-frequency details while maintaining structural consistency. Furthermore, the significant inference latency advantage confirms its practical value for real-world deployment.

## 6. Conclusions

This paper has proposed FreeTransformSR, an efficient network for lightweight image super-resolution. The method centers on a channel-wise free low-rank learnable transform that replaces fixed frequency-domain bases in a data-driven manner, achieving adaptive feature modulation with low parameter overhead. Building upon this, the local feature modulation branch complements high-frequency details through depthwise convolution, working synergistically with transform-domain processing. The soft complexity adaptive module dynamically fuses the outputs of local convolution and window self-attention branches via a lightweight gating network, adaptively adjusting the fusion ratio according to regional characteristics. The adaptive intensity modulation further adjusts transform-domain response strength at the sample level based on input features.

Experiments on five benchmark datasets demonstrate that FreeTransformSR achieves competitive reconstruction accuracy with significantly fewer parameters and FLOPs. Specifically, the ×2 model achieves 32.41 dB on BSD100 with only 634K parameters, and the ×4 model achieves 27.00 dB on Urban100, with inference speed surpassing competing methods. Ablation studies validate the effectiveness of each module.

Future work will explore extending the framework to complex real-world degradation scenarios and larger-scale training data, as well as investigating adaptive window mechanisms to better capture multi-scale structural information. FreeTransformSR provides an efficient and practical solution for super-resolution deployment in resource-constrained scenarios.